\documentclass[]{bytedance_seed}

\usepackage[toc,page,header]{appendix}

\usepackage{natbib}

\usepackage{CJKutf8}

\usepackage{xargs}  

\usepackage{todonotes}  

\usepackage{multirow}

\usepackage{cleveref}

\usepackage{amsmath}
\usepackage{dsfont}

\usepackage{subcaption}

\usepackage{svg}

\usepackage{mathrsfs}
\usepackage{adjustbox}
\usepackage{multirow}
\usepackage{multicol}
\usepackage{tcolorbox}
\usepackage{changepage}
\usepackage{enumitem}
\usepackage{graphicx}
\usepackage{amssymb}
\usepackage{xcolor}
\usepackage{tabularx} 
\usepackage{makecell} 
\usepackage[dvipsnames]{xcolor}

\usepackage{minitoc}

\title{UniGRPO: Unified Policy Optimization for Reasoning-Driven Visual Generation}

\author{
  Jie Liu\texorpdfstring{$^{1,2\ast}$}{} \hspace{0.1cm}
  Zilyu Ye\texorpdfstring{$^{2\ast}$}{} \hspace{0.1cm}
  Linxiao Yuan\texorpdfstring{$^{2}$}{} \hspace{0.1cm}
  Shenhan Zhu\texorpdfstring{$^{2}$}{}  \hspace{0.1cm} 
  Yu Gao\texorpdfstring{$^{2}$}{} \hspace{0.1cm} 
  Jie Wu\texorpdfstring{$^{2\ddagger}$}{} \\[0.1cm]\hspace{0.1cm}
  Kunchang Li\texorpdfstring{$^{2}$}{} \hspace{0.1cm}
  Xionghui Wang\texorpdfstring{$^{2}$}{} \hspace{0.1cm}
  Xiaonan Nie\texorpdfstring{$^{2}$}{} \hspace{0.1cm}
  Weilin Huang\texorpdfstring{$^{2\S}$}{}\hspace{0.1cm}
  Wanli Ouyang\texorpdfstring{$^{1}$}{}
}

\affiliation[1]{The Chinese University of Hong Kong}
\affiliation[2]{ByteDance Seed}
\contribution[*]{Equal contribution} 
\contribution[\ddagger]{Project lead}
\contribution[\S]{Corresponding author} 

\begin{document}
\begin{CJK*}{UTF8}{gbsn}

\abstract{
Unified models capable of interleaved generation have emerged as a promising paradigm, with the community increasingly converging on autoregressive modeling for text and flow matching for image generation. To advance this direction, we propose a unified reinforcement learning framework tailored for interleaved generation. We validate our approach on its fundamental unit: a single round of reasoning-driven image generation, where the model first expands the user prompt through reasoning, followed by image synthesis.
Formulating this multimodal generation process as a Markov Decision Process with sparse terminal rewards, we introduce UniGRPO to jointly optimize text and image generation policies using GRPO. Adopting a minimalist methodology to avoid over-design, we leverage established training recipes for both modalities by seamlessly integrating standard GRPO for reasoning and FlowGRPO for visual synthesis. To ensure scalability to multi-round interleaved generation, we introduce two critical modifications to the original FlowGRPO: (1) eliminating classifier-free guidance to maintain linear, unbranched rollouts, which is essential for scaling to complex scenarios involving multi-turn interactions and multi-condition generation (e.g., editing); and (2) replacing the standard latent KL penalty with an MSE penalty directly on the velocity fields, providing a more robust and direct regularization signal to mitigate reward hacking effectively. Our experiments demonstrate that this unified training recipe significantly enhances image generation quality through reasoning, providing a robust and scalable baseline for the future post-training of fully interleaved models.
}

\maketitle

\section{Introduction}

\begin{figure}[!ht]
\centering
\includegraphics[width=0.95\textwidth]{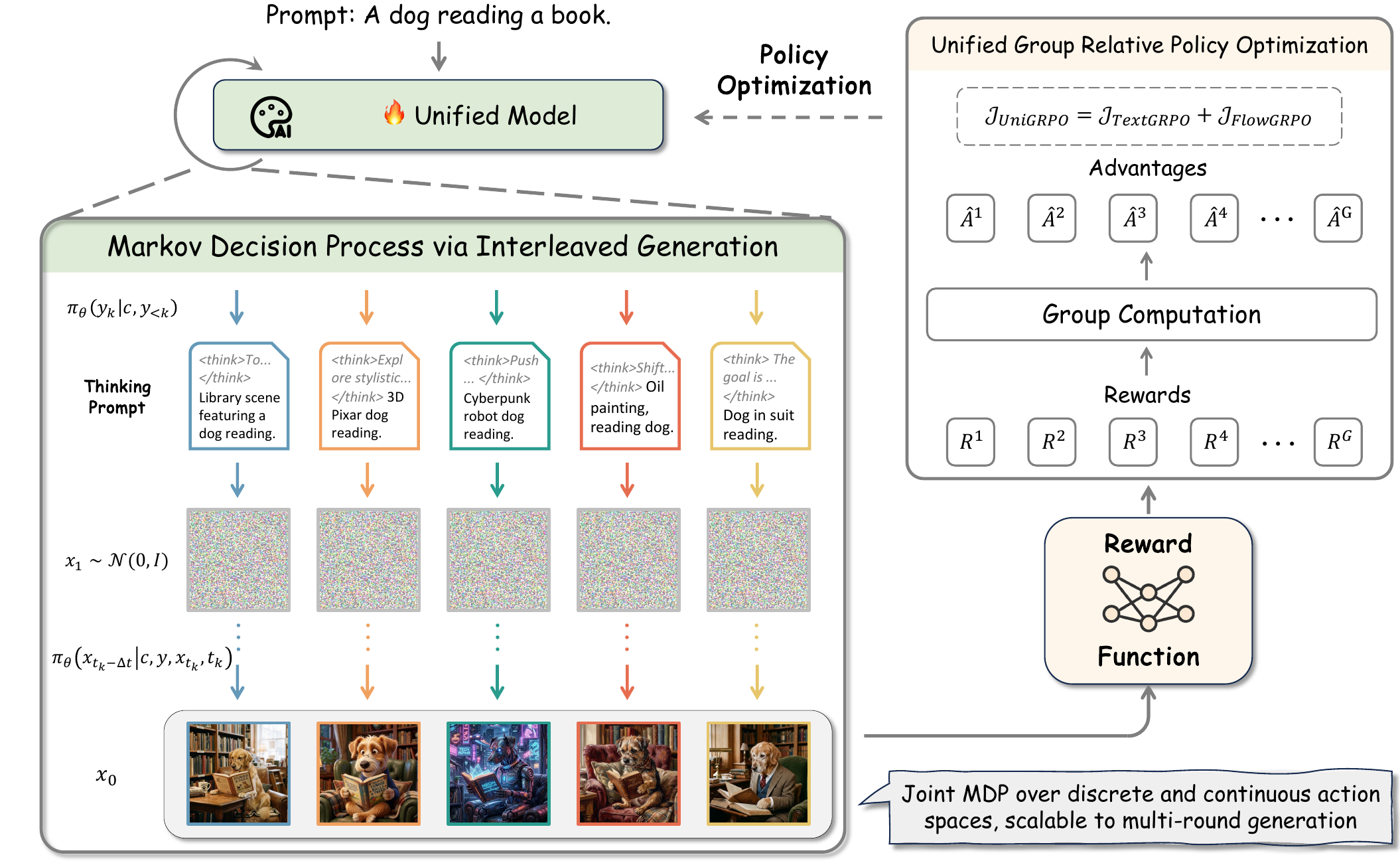}
\caption{\textbf{Overview of UniGRPO.} By formulating interleaved generation as a joint MDP, this illustration demonstrates how UniGRPO jointly optimizes discrete language actions ($y_k$) in the LLM's next-token prediction, and continuous visual actions ($x_{t_k-\Delta t}$) in flow matching. Both policies are updated using group-relative advantages derived from sparse terminal rewards.}
\vspace{-4mm}
\label{fig:teaser}
\end{figure}

\begin{figure}[!ht]
\centering
\includegraphics[width=0.87\textwidth]{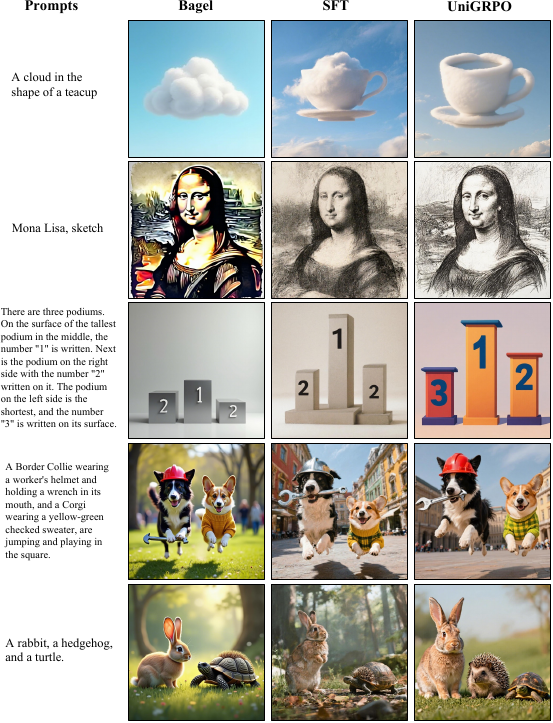}
\vspace{-2mm}
\caption{T2I qualitative comparison.}
\vspace{-6mm}
\label{fig:t2i_quality}
\end{figure}

The evolution of generative AI is rapidly progressing toward unified multimodal models~\cite{deng2025emerging,wei2025skywork,zhao2025unified,xie2024show,zhou2024transfusion} capable of interleaved generation~\cite{liao2025mogao}. A pivotal advantage of this emerging paradigm is the potential to effectively leverage test-time compute through iterative reasoning — refining prompts, generating images, and reflecting on outputs across multiple rounds to tackle complex image synthesis tasks~\cite{huang2025interleaving}. As the boundaries between modalities blur, the community is increasingly gravitating toward a robust architectural synergy: Autoregressive (AR)~\cite{radford2019language} models for text generation paired with Flow Matching~\cite{lipman2022flow, rectified_flow} for visual synthesis~\cite{deng2025emerging,xie2024show,zhou2024transfusion,liao2025mogao}. This combination harnesses the reasoning capabilities of Large Language Models (LLMs) alongside the high-fidelity generation strengths of Flow-based models.

In this work, we argue that advancing interleaved generation requires a unified Reinforcement Learning (RL) framework that jointly optimizes text and image generation policies. Rather than immediately scaling to long-horizon multi-turn generation, we validate our framework on its fundamental unit: a single round of reasoning-driven image generation. This setting already encompasses both text and image generation, covering the essential components of interleaved generation. In the absence of open-source base models natively capable of full interleaved generation, it serves as a meaningful and principled testbed for validating our unified RL framework.

To this end, we propose UniGRPO, a unified RL framework formulating the entire "Prompt $\rightarrow$ Thinking $\rightarrow$ Image" sequence as a single Markov Decision Process (MDP)~\cite{puterman1990markov}. Adopting a minimalist methodology to avoid over-design, we integrate established training recipes for both modalities: standard GRPO~\cite{grpo} for the reasoning component and FlowGRPO~\cite{liu2025flow} for visual synthesis. Under sparse terminal rewards, UniGRPO jointly optimizes both text and image generation policies, encouraging the model to produce more informative reasoning texts while simultaneously improving the visual synthesis process itself.

Crucially, our design choices are driven by the goal of scalability to future multi-round and multi-condition scenarios (e.g., complex editing tasks). We introduce two critical modifications to the standard Flow Matching RL training recipe within our framework. First, we eliminate Classifier-Free Guidance (CFG)~\cite{ho2022classifier} during training. While CFG is a standard inference technique, its removal ensures that the generation process remains a linear, unbranched rollout, which is essential for scaling to complex scenarios involving multi-turn interactions and multi-condition generation. Second, we replace the standard latent KL penalty with an MSE penalty directly on the velocity fields. This provides a more robust and direct regularization signal that effectively mitigates reward hacking, ensuring the optimization remains well-grounded.
Our contributions can be summarized as follows:
\begin{itemize}
\item Unified RL Framework for Reasoning-Driven Image Generation: We propose UniGRPO, a minimalist framework that formulates the Prompt $\rightarrow$ Thinking $\rightarrow$ Image sequence as a single MDP, jointly optimizing AR text and flow-matching image policies. We validate this framework on the fundamental unit of interleaved generation, demonstrating that jointly optimizing reasoning and visual synthesis improves image generation quality.

\item Scalable Flow Matching RL Adaptations: We introduce two critical modifications to FlowGRPO: eliminating CFG to ensure unbranched rollouts, and replacing the standard latent KL penalty with an MSE penalty directly on the velocity fields for more robust reward hacking mitigation. Together, these adaptations are essential for scaling to multi-turn and multi-condition generation scenarios.

\item We demonstrate that our unified training recipe effectively optimizes the model under sparse terminal rewards, establishing a robust and scalable baseline for future post-training of fully interleaved models.
\end{itemize}
\section{Related Work}

\subsection{RL for LLMs}
Recent LLM advancements rely on Reinforcement Learning (RL) for alignment and reasoning. While PPO~\cite{schulman2017proximal} is a standard approach, the highly efficient Group Relative Policy Optimization (GRPO)~\cite{grpo} eliminates the value model by using group-relative baselines. This efficiency drives reasoning-intensive models using Chain-of-Thought (CoT)~\cite{wei2022chain}, such as DeepSeek-R1. Our work adapts GRPO to efficiently optimize the intermediate "thinking" tokens prior to visual synthesis.

\subsection{RL for Diffusion and Flow Matching Models}
Aligning text-to-image (T2I) models with human intent has been extensively explored, primarily through reward-driven optimization~\cite{prabhudesai2023aligning, clark2023directly, xu2024imagereward, prabhudesai2024video} and Reward Weighted Regression (RWR)~\cite{peng2019advantage, fan2025online, lee2023aligning,dong2023raft}. Currently, Direct Preference Optimization (DPO)~\cite{rafailov2024direct, wallace2024diffusion, videoalign, yang2024using, liang2024step, yuan2024self, liu2024videodpo, zhang2024onlinevpo, furuta2024improving,liang2025aesthetic} and PPO-style policy gradients~\cite{schulman2017proximal, black2023training, fan2024reinforcement, gupta2025simple, miao2024training, zhao2025score} have become standard frameworks for fine-tuning diffusion models, alongside various training-free guidance methods~\cite{yeh2024training, tang2024tuning, song2023loss}. However, adapting these established RL paradigms to the deterministic ODEs of modern flow matching architectures requires specific stochastic formulations. To address this, FlowGRPO~\cite{liu2025flow} and DanceGRPO~\cite{xue2025dancegrpo} introduce a method to apply policy gradients to flow models by reformulating the generation process into a stochastic SDE. Subsequently, several works~\cite{densegrpo,diffusionnft,huang2025mix,gdro,egrpohighentropy,gardoreinforcingdiffusion,cps,g2rpogranulargrpo,neighborgrpocontrastive,finetuningflowmatching,yang2025tempflowgrpowhentiming,luoThu,yu2025smartgrposmartlysampling,liu2025inferencetimealignmentcontrol} have further improved upon FlowGRPO by enhancing training stability, reward design, or sample efficiency. Building on this line of work, our work extends the RL framework to jointly optimize both language reasoning and visual synthesis.

\subsection{Unified Multimodal Understanding and Generation Models}
Multimodal understanding and image generation have long evolved independently, with
autoregressive models dominating the former and diffusion models the latter.
Recent work seeks to unify both capabilities within a single framework.
One line of research applies vector quantization to visual signals so that image and text tokens
share a unified autoregressive training space, as in
Chameleon~\cite{team2024chameleon}, Emu3~\cite{wang2024emu3}, and VILA-U~\cite{wu2024vila}.
Another line combines autoregressive and diffusion objectives:
Show-o~\cite{xie2024show} and Transfusion~\cite{zhou2024transfusion} train a single transformer
with mixed next-token prediction and diffusion losses, while
Bagel~\cite{deng2025emerging} and Mogao~\cite{liao2025mogao} further scale this hybrid paradigm with
large-scale interleaved multimodal data, demonstrating strong emerging capabilities in
complex reasoning and coherent interleaved text-image generation.
As surveyed by Zhang et al.~\cite{zhao2025unified}, key challenges remain in tokenization
strategy, cross-modal attention design, and training data construction.

\subsection{Concurrent Work}

Concurrent with our work, several studies independently apply RL to unified or
joint multimodal generation.
R3~\cite{ye2026understanding} proposes a generate-understand-regenerate loop to mitigate the
understanding-generation trade-off, but validates on benchmark-specific prompts rather
than general-purpose training.
DualGRPO~\cite{kou2026think} jointly optimizes a separate LLM model and diffusion backbone
via a tree-structured rollout, yet this design is incompatible with true interleaved
multimodal generation.
PromptRL~\cite{wang2026promptrl} similarly trains disjoint language and flow models in a joint
RL loop, but on limited training datasets.
SepGRPO~\cite{jiao2025thinkgen} is also built on BAGEL and proposes alternating RL between
the MLLM and DiT modules, but the two components are trained separately rather than
jointly optimized end-to-end.
In contrast, our method is built on a single unified model, trained with general-purpose
prompts at 1024 resolution, with a scalable algorithm design built upon an improved
FlowGRPO. We further provide comprehensive comparisons against a wide range of diffusion
RL baselines, yielding broader and more robust performance gains across diverse benchmarks.
\section{Preliminary}
\label{sec:preliminary}

In this section, we establish the theoretical foundations for optimizing generative policies using Unified Group Relative Policy Optimization (UniGRPO), covering both discrete text generation and continuous flow-based visual generation.

\subsection{Text GRPO}
For the autoregressive text component, we adopt the standard GRPO~\cite{grpo} formulation. Given a prompt $c$, the policy $\pi_{\theta}$ generates a group of $G$ outputs $\{y_i\}_{i=1}^G$. The optimization objective maximizes the expected reward while constraining the policy update via importance sampling clipping.

The advantage for the $i$-th sample is computed relatively within the group:
\begin{equation}
    \hat{A}_i = \frac{R_i - \text{mean}(\{R_j\}_{j=1}^G)}{\text{std}(\{R_j\}_{j=1}^G)}.
\end{equation}
The objective function is defined as:
\begin{equation}
    \mathcal{J}_{\text{Text}}(\theta) = \frac{1}{G} \sum_{i=1}^G \frac{1}{|y_i|} \sum_{k=1}^{|y_i|} \left( \min \left( r_{i,k}\, \hat{A}_i,\; \text{clip}(r_{i,k}, 1{-}\epsilon, 1{+}\epsilon)\, \hat{A}_i \right) - \beta_{\text{txt}}\, D_{\text{KL}}(\pi_\theta \| \pi_{\text{ref}}) \right),
\end{equation}
where $r_{i,k} = \frac{\pi_{\theta}(y_{i,k} \mid y_{i,<k})}{\pi_{\theta_{\text{old}}}(y_{i,k} \mid y_{i,<k})}$ denotes the importance ratio at step $k$.

\subsection{Flow GRPO}
For the visual component, we utilize FlowGRPO~\cite{liu2025flow}, which adapts reinforcement learning to flow matching models by converting the deterministic Ordinary Differential Equation (ODE) into a Stochastic Differential Equation (SDE) to enable exploration.

\paragraph{SDE Sampling.} To introduce the necessary stochasticity for RL exploration, the sampling process is formulated as:
\begin{equation}
    \Delta x_{t_k} = \left[v_{\theta}(x_{t_k}, t_k) + \frac{\sigma_{t_k}^2}{2\,t_k}\bigl(x_{t_k} + (1-t_k)\,v_{\theta}(x_{t_k}, t_k)\bigr)\right]\Delta t + \sigma_{t_k} \sqrt{\Delta t}\;\epsilon,
\end{equation}
where $\sigma_{t_k}$ controls the noise level and $\epsilon \sim \mathcal{N}(0, I)$. For training efficiency, we adopt the FlowGRPO-Fast variant~\cite{liu2025flow}, which employs a hybrid sampling strategy. Specifically, denoising steps within a continuous time window are performed via SDE and optimized with gradient tracking, while the remaining steps follow standard ODE sampling without gradient computation. This significantly reduces computational overhead while preserving optimization effectiveness.

\paragraph{Mitigating Reward Hacking via RatioNorm.} Standard importance-ratio clipping often fails in diffusion models because the distribution of importance ratios 
\begin{equation}
r_{t_k}(\theta) = \frac{p_\theta(x_{t_k-\Delta t} \mid c, y, x_{t_k}, t_k)}{p_{\theta_{\text{old}}}(x_{t_k-\Delta t} \mid c, y, x_{t_k}, t_k)},
\end{equation}
is systematically left-shifted (mean $< 1$) and exhibits inconsistent variance across timesteps~\cite{wang2025grpo}. This prevents the clipping mechanism from constraining overconfident positive updates, leading to severe reward hacking. To address this, we adopt the Ratio Normalization (RatioNorm) proposed in GRPO-Guard~\cite{wang2025grpo}. This method standardizes the log-importance ratio to center its distribution around zero, thereby restoring the effectiveness of the clipping bounds:
\begin{equation}
    \log \tilde{r}_{t_k}(\theta) = \sigma_{t_k} \sqrt{\Delta t} \left( \log r_{t_k}(\theta) + \frac{\|\Delta \mu_{\theta}(x_{t_k}, t_k)\|^2}{2\,\sigma_{t_k}^2\, \Delta t} \right),
\end{equation}
where $\Delta\mu_\theta(x_{t_k}, t_k) \triangleq \mu_{\theta_{\text{old}}}(x_{t_k}, t_k) - \mu_\theta(x_{t_k}, t_k)$ is the mean drift between the current and reference policies.

Combining the hybrid SDE sampling strategy with the RatioNorm mechanism, the final FlowGRPO objective is computed exclusively over the SDE timestep subset $\mathcal{T}_{\text{SDE}}$:
\begin{equation}
    \mathcal{J}_{\text{Flow}}(\theta) = \frac{1}{G} \sum_{i=1}^G \frac{1}{|\mathcal{T}_{\text{SDE}}|} \sum_{t_k \in \mathcal{T}_{\text{SDE}}} \left( \min \left( \tilde{r}_{i,t_k}\, \hat{A}_i,\; \text{clip}(\tilde{r}_{i,t_k}, 1{-}\epsilon, 1{+}\epsilon)\, \hat{A}_i \right) - \beta_{\text{img}}\, D_{\text{KL}}(\pi_\theta \| \pi_{\text{ref}})\right),
\end{equation}
where $|\mathcal{T}_{\text{SDE}}|$ denotes the number of denoising steps within the continuous SDE window.

\section{Method}
\label{sec:method}

Building upon these foundations, we propose \textbf{UniGRPO}, a unified framework that jointly optimizes multimodal generation policies within a single reinforcement learning loop.

\subsection{Multimodal Generation as a Markov Decision Process}
We formulate interleaved generation as a sequential MDP
$(\mathcal{S}, \mathcal{A}, P, R)$, where each MDP step $k$
corresponds to a single token prediction during the text phase
and a single denoising step during the image phase.

\begin{itemize}
    \item \textbf{State Space $\mathcal{S}$:}
    The state evolves through two phases.
    In the \emph{text phase}, $s^{\text{txt}}_k = (c,\; y_{<k})$
    comprises the input prompt $c$ and all previously generated
    reasoning tokens $y_{<k}$.
    In the \emph{image phase},
    $s^{\text{img}}_k = (c,\; y,\; x_{t_k},\; t_k)$
    includes the prompt, the completed reasoning trace $y$,
    the noisy image latent $x_{t_k}$, and the current flow
    time $t_k$.

    \item \textbf{Action Space $\mathcal{A}$:}
    In the text phase, $a^{\text{txt}}_k \in \mathcal{V}$
    is a single token drawn from the vocabulary.
    In the image phase,
    $a^{\text{img}}_k = x_{t_k - \Delta t} \in \mathbb{R}^d$
    is the denoised latent at the next flow step.

    \item \textbf{Transition $P$:}
    Both phases are deterministic given the action:
    the text transition appends $a^{\text{txt}}_k$ to the
    token sequence, while the image transition advances
    the latent from $x_{t_k}$ to $x_{t_k - \Delta t}$.

    \item \textbf{Reward $R$:}
    A sparse terminal reward $R(x_0, c)$ is assigned
    only after the image latent has been fully denoised
    to $x_0$; all intermediate steps receive zero reward.
\end{itemize}

\subsection{UniGRPO Framework}
Given a unified model $\pi_\theta$ that performs interleaved generation, UniGRPO models the entire generation process as a MDP and optimizes it through group relative policy optimization. Specifically, for a given prompt $c$, we first sample $G$ reasoning chains $\{y_i\}_{i=1}^G$ via $\pi_\theta(a_k^{\text{txt}} \mid s_{k}^{\text{txt}})$. Each reasoning chain then conditions the same model to generate a corresponding image trajectory $\{x_i\}_{i=1}^G$ via $\pi_\theta(a_k^{\text{img}} \mid s_{k}^{\text{img}})$ with a hybrid SDE-ODE integrator. We compute group-relative advantages $\hat{A}_i$ based on the terminal rewards of the completed multimodal trajectories. These advantages are used to update $\pi_\theta$ through a unified objective:
\begin{equation}
\mathcal{J} = \mathcal{J}_{\text{Text}} + \lambda\, \mathcal{J}_{\text{Flow}},
\end{equation}
where $\lambda$ is a hyperparameter controlling the relative weight of the image generation objective. To equally balance the reasoning and synthesis tasks, we simply set $\lambda = 1$ across all our experiments.
To ensure scalability to multi-round interleaved generation, we introduce two critical modifications to the training recipe.

\paragraph{Eliminating Classifier-Free Guidance.}
Standard flow matching inference typically relies on CFG to enhance prompt adherence, requiring two model evaluations per step (conditional and unconditional). Crucially, this computational burden scales with the number of conditions; for multi-condition generation such as image editing, CFG demands at least three evaluations per step. Furthermore, this complexity compounds in multi-round interleaved generation, where the system must continuously manage and branch multiple conditional contexts across alternating text and image phases. In an RL setting, this multiplication of function evaluations and context branches drastically inflates computational and memory costs, while creating a branched computation graph that severely complicates gradient estimation. We therefore train UniGRPO entirely without CFG, enforcing a linear, unbranched rollout. While removing CFG typically degrades prompt adherence, our framework compensates for this during training. By explicitly maximizing the expected reward—which evaluates text-image alignment and visual quality—we internalize the alignment capabilities directly into the policy weights. This establishes a highly efficient pipeline that naturally scales to complex multi-condition, multi-round interaction generation.

\paragraph{Velocity-Based Regularization.}
Preventing reward hacking is a primary challenge in RL for visual generation. In the above SDE formulation, the step-wise transition probabilities are Gaussian, meaning the exact local KL divergence in the latent space can be analytically computed. Specifically, this exact KL evaluates to the squared difference in predicted velocities, weighted by the inverse noise variance ($1/\sigma_{t_k}^2$). However, this inherent weighting applies an uneven penalty across the generative trajectory. For instance, at timesteps with high noise variance, the KL penalty becomes excessively small. This inconsistency creates temporal vulnerabilities that the RL optimizer can easily exploit. To achieve a more robust and consistent constraint, we drop this timestep-dependent weighting and apply a Mean Squared Error (MSE) penalty directly on the unweighted velocity fields:
\begin{equation}
    \mathcal{L}_{\text{MSE}}(\theta) = \| v_{\theta}(x_{t_k}, t_k, y) - v_{\text{ref}}(x_{t_k}, t_k, y) \|^2.
\end{equation}
This unweighted formulation explicitly forces the RL-tuned vector field to remain close to the pre-trained reference model uniformly across all noise levels. Empirically, we find that this uniform regularization leaves fewer loopholes for policy exploitation, proving significantly more effective at mitigating reward hacking while safely preserving the base model's generative priors.
\section{Experiments}
\label{sec:experiments}

This section presents the empirical validation of the proposed UniGRPO framework. We begin by outlining the experimental setup—including the pretrained model, reward formulation, baselines, and evaluation protocols. Detailed hyperparameter settings are deferred to Appendix~\ref{app:tab:hyperparameters}. Following this, we compare UniGRPO against strong baselines and conclude with ablation studies to evaluate critical design choices.

\subsection{Experimental Settings}

\paragraph{The Pretrained Model.}
As a preliminary exploration into reinforcement learning for interleaved generation, we require a backbone capable of handling mixed-modal outputs. We adopt Bagel~\cite{deng2025emerging}, a model architecture with inherent interleaved generation potential. However, we observed that the vanilla Bagel exhibits limited instruction-following capabilities and suboptimal image generation quality. To establish a strong baseline, we performed Supervised Fine-Tuning (SFT) on Bagel using a curated internal dataset. This process significantly boosted performance (see Table~\ref{tab:main_results}). Unless otherwise stated, all subsequent baselines and experiments utilize this finetuned Bagel as the starting checkpoint.

\paragraph{Reward Model.}
A key advantage of the GRPO algorithm is its flexibility; it does not require differentiable reward functions, allowing the integration of black-box verifiers or VLM-based feedback. However, to ensure a fair comparison with gradient-based baselines like ReFL~\cite{xu2024imagereward} (which necessitates differentiable rewards), we utilize a differentiable reward formulation for the main experiments. Specifically, we employ the exact same reward model as utilized in RewardDance~\cite{wu2025rewarddance}. This model is fine-tuned based on InternVL~\cite{chen2024internvl} using collected user preference data, explicitly designed to measure the consistency between generated images and user prompts. It is important to note that while ReFL is restricted to such differentiable objectives, UniGRPO is compatible with a broader range of verifier-based rewards.

\begin{table*}[t]
\small
\centering
\setlength{\tabcolsep}{8pt}
\caption{\textbf{Main results on TA and GenEval.} All RL methods start from the Bagel checkpoint after SFT. ``Thinking'' denotes whether the method explicitly generates intermediate reasoning tokens. $-$ indicates training collapse.}
\vspace{-2mm}
\label{tab:main_results}
\begin{tabular}{lccc}
\toprule
\textbf{Model / Method} & \textbf{Thinking} & \textbf{TA Score} & \textbf{GenEval} \\
\midrule
Bagel & $\times$ & 0.6810 & 0.78 \\
Bagel & $\checkmark$ & 0.7132 & 0.79 \\
\midrule
SFT & $\times$ & 0.7486 & 0.83 \\
SFT & $\checkmark$ & 0.7769 & 0.82 \\
\midrule
ReFL & $\times$ & 0.7786 & 0.85 \\
ReFL & $\checkmark$ & 0.8120 & 0.84 \\
FPO & $\times$ & 0.7893 & 0.87 \\
FPO & $\checkmark$ & 0.8159 & 0.85 \\
FlowGRPO & $\times$ & 0.8112 & 0.88 \\
FlowGRPO & $\checkmark$ & 0.8208 & 0.86 \\
TextGRPO & $\checkmark$ & 0.8078 & 0.88 \\
\midrule
ReFL (w/ Thinking) & $\checkmark$ & 0.7804 & 0.83 \\
ReFL (w/ Thinking) + TextGRPO & $\checkmark$ & 0.7987 & 0.87 \\
UniFPO & $\checkmark$ & $-$ & $-$ \\
\textbf{UniGRPO (Ours)} & $\checkmark$ & \textbf{0.8381} & \textbf{0.90} \\
\bottomrule
\end{tabular}
\end{table*}
\vspace{-2mm}
\paragraph{Baselines.}
\textbf{ReFL} directly fine-tunes diffusion models by viewing reward model scores as human preference losses and back-propagating gradients to a randomly-picked late timestep $t$.
\textbf{ReFL w/ Thinking} generates thinking prompts during training and optimizing only the image generation part using the ReFL objective.
\textbf{ReFL + TextGRPO} follows a two-stage paradigm: initializing from the trained ReFL w/ Thinking checkpoint and subsequently optimizing the text generation module using TextGRPO.
\textbf{FPO / AWR}~\cite{mcallister2025flow, xue2025advantage} serves as an alternative to FlowGRPO. Unlike FlowGRPO which introduces SDE perturbations for exploration, FPO utilizes the forward process to obtain $x_t$ and uses the Evidence Lower Bound (ELBO) of the denoising process as a surrogate for $\log p_\theta(x_0|c)$ to compute importance sampling weights.
\textbf{UniFPO} denotes a unified framework analogous to UniGRPO, where the text component is optimized via TextGRPO and the image synthesis component is trained using the FPO objective.

\paragraph{Evaluation Metrics.}
We employ two benchmarks to evaluate generation quality and prompt alignment:
\begin{itemize}
    \item Text Alignment (TA) Benchmark: Our internal evaluation set consisting of 150 diverse prompts. For each prompt, we generate 4 images. Evaluation is performed by a VLM, which assesses the outputs against multiple specific exam points defined for each prompt. Each exam point receives a binary score (1 for correct, 0 for incorrect), and the score for a single image is calculated as the average score across all its associated exam points. The final reported metric is the overall average score across all evaluated images. We refer to RewardDance~\cite{wu2025rewarddance} for further details on this scoring mechanism.
    \item GenEval~\cite{geneval}: A standard benchmark assessing Text-to-Image models on complex compositional capabilities, including object counting, spatial relations, and attribute binding.
\end{itemize}
\subsection{Main Results}
\begin{figure}
\centering
\vspace{-4mm}
\includegraphics[width=0.48\textwidth]{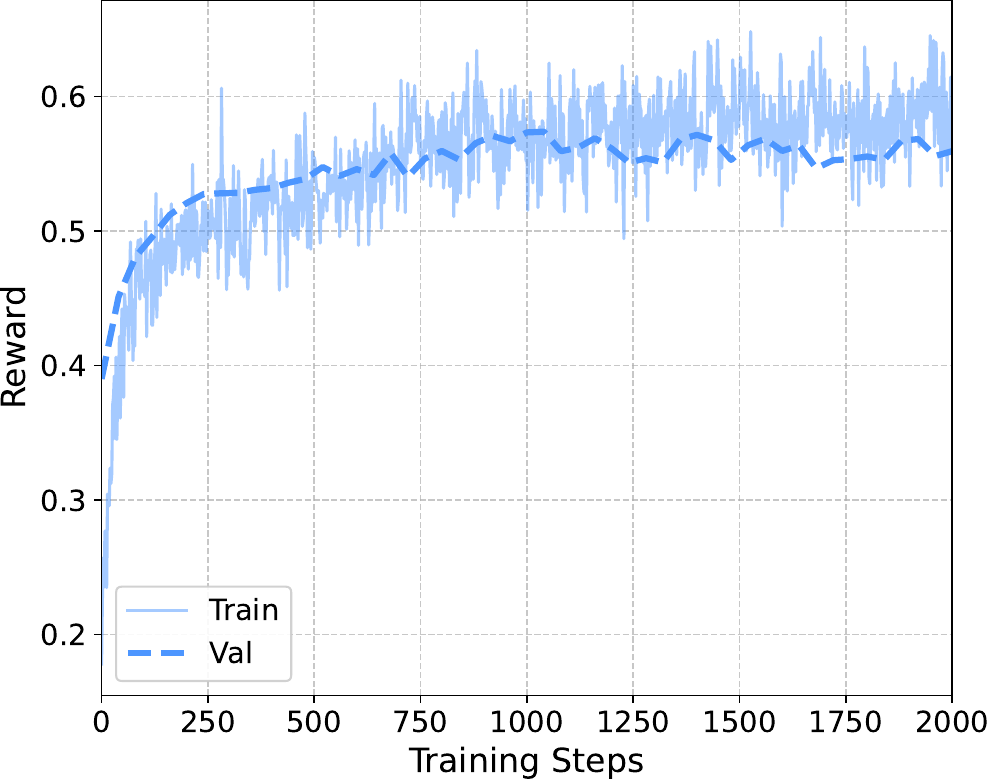}
\caption{Training and Validation reward curves of UniGRPO on the Finetuned Bagel base model at a resolution of 1024. The x-axis represents the gradient update steps.}
\vspace{-4mm}
\label{fig:unigrpo_curve}
\end{figure}

We begin by analyzing the learning dynamics of UniGRPO, presenting the training and validation reward curves in Figure~\ref{fig:unigrpo_curve} alongside qualitative generation examples in Figure~\ref{fig:t2i_quality}. Next, we benchmark our framework against several established baselines: ReFL, FPO, FlowGRPO, TextGRPO, and hybrid approaches. The quantitative comparisons are summarized in Table~\ref{tab:main_results}.
\paragraph{Benchmark Performance.}
The results in Table~\ref{tab:main_results} indicate that SFT significantly improves the base capabilities of Bagel. Among the RL methods, UniGRPO achieves state-of-the-art performance, scoring 0.8381 on TA and 0.90 on GenEval. Notably, UniFPO failed to converge, underscoring the stability advantages of our GRPO-based formulation. The comparison between UniGRPO, FlowGRPO, and TextGRPO confirms that jointly optimizing both the reasoning and synthesis policies yields gains superior to optimizing either component in isolation. Furthermore, we observe that enabling the explicit reasoning chain ("Thinking") on the Bagel model does not consistently improve GenEval scores. As noted by the Bagel authors, their reasoning module is primarily trained for knowledge-based reasoning and may not be ideally suited for short prompt rewriting tasks found in benchmarks like GenEval \footnote{\small See \url{https://github.com/ByteDance-Seed/Bagel/issues/109\#issuecomment-2934226809}}. However, our UniGRPO framework successfully leverages the reasoning chain to achieve SOTA performance. Please see Appendix~\ref{app:subsec:detailed-geneval} for the detailed metrics of GenEval.

\paragraph{Qualitative Analysis.}
As illustrated in Figure~\ref{fig:t2i_quality}, the original Bagel tends to generate images with oversaturated colors and noticeable synthetic artifacts. While SFT helps mitigate these synthetic artifacts, it compromises image sharpness, resulting in noticeable blurriness upon close inspection. Overcoming this limitation, our proposed UniGRPO significantly enhances both aesthetic quality and text-image alignment, yielding photorealistic, finely detailed images that faithfully reflect complex user prompts. Beyond visual quality, we analyze the models' internal reasoning processes in Figures~\ref{fig:unigrpo_think_img}, \ref{fig:bagel_text_img} $\&$ \ref{fig:sft_text_img}. Note that during SFT, we standardized Bagel's original \texttt{<think>} format to align with prevalent LLM conventions. While the base Bagel and SFT models generate detailed reasoning texts, these traces can sometimes lose focus or become disconnected from the core visual generation task. In contrast, UniGRPO optimizes the reasoning phase to be highly purposeful and task-oriented. By explicitly aligning the thought process with the final visual reward, UniGRPO produces reasoning traces that tightly couple with and effectively guide the subsequent image synthesis.

\subsection{Ablation Study}

We conduct ablation studies to validate our specific architectural and training decisions, including the removal of CFG and the choice of KL regularization.

\begin{figure}[h]
\centering
\includegraphics[width=0.5\textwidth]{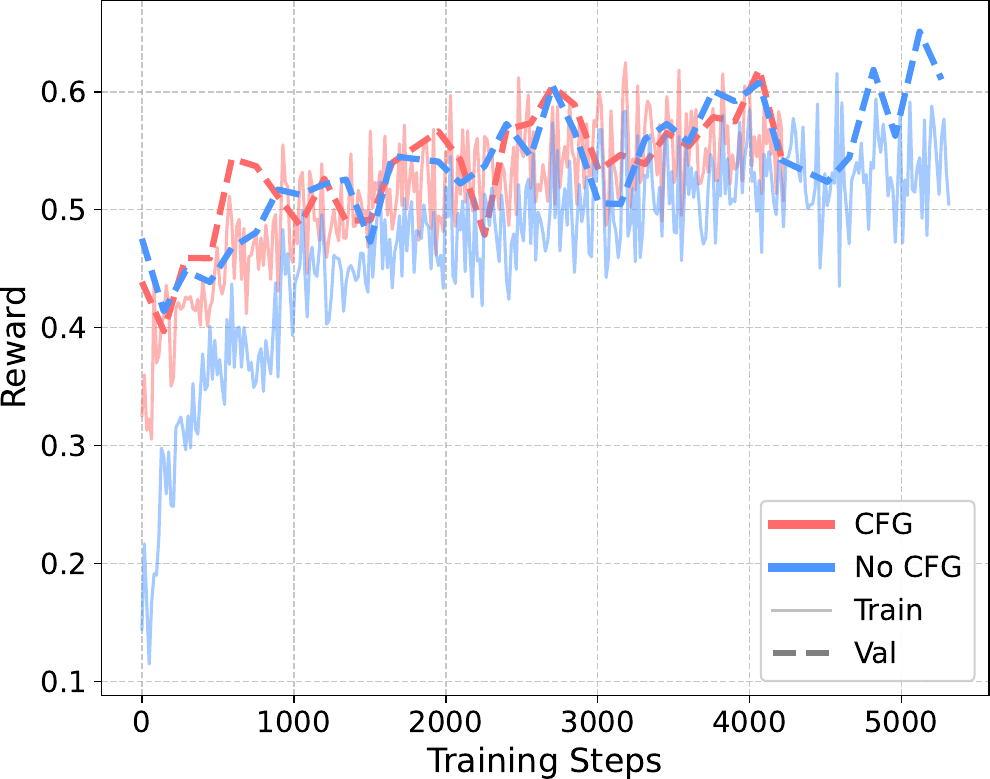}
\caption{\textbf{Ablation Study on CFG.} Removing CFG during training yields comparable or superior performance, showing that CFG is unnecessary for RL-based training. Note that CFG is applied at evaluation for all settings. Furthermore, these results are not directly comparable to the curves in Figure \ref{fig:unigrpo_curve}, as this ablation uses the original Bagel as the base model at a resolution of 512.}
\vspace{-2mm}
\label{fig:cfg_train}
\end{figure}

\paragraph{Impact of CFG-Free Training.}
We compared training UniGRPO with and without CFG. As shown in Figure~\ref{fig:cfg_train}, although CFG during training yields images with higher rewards, removing CFG results in comparable or better final performance when evaluated with CFG. This confirms that CFG is unnecessary for RL-based alignment, rendering the computationally expensive branched CFG rollouts unnecessary during training.

\paragraph{Regularization Strategies.}
Preventing reward hacking is critical in RL. We compared three strategies: (1) No KL, (2) Latent KL (standard practice), and (3) Velocity MSE (Ours). As shown in Figure~\ref{fig:ablattion_kl}, removing KL leads to reward hacking where metrics are high but quality degrades. Velocity MSE achieves the best balance, constraining the vector field to the reference model while maintaining strong generation performance.
\definecolor{pltnokl}{HTML}{7ac588}
\definecolor{pltkl}{HTML}{6da2ee}
\definecolor{pltmse}{HTML}{f2736e}
\begin{figure}[t]
\centering
\includegraphics[width=1\textwidth]{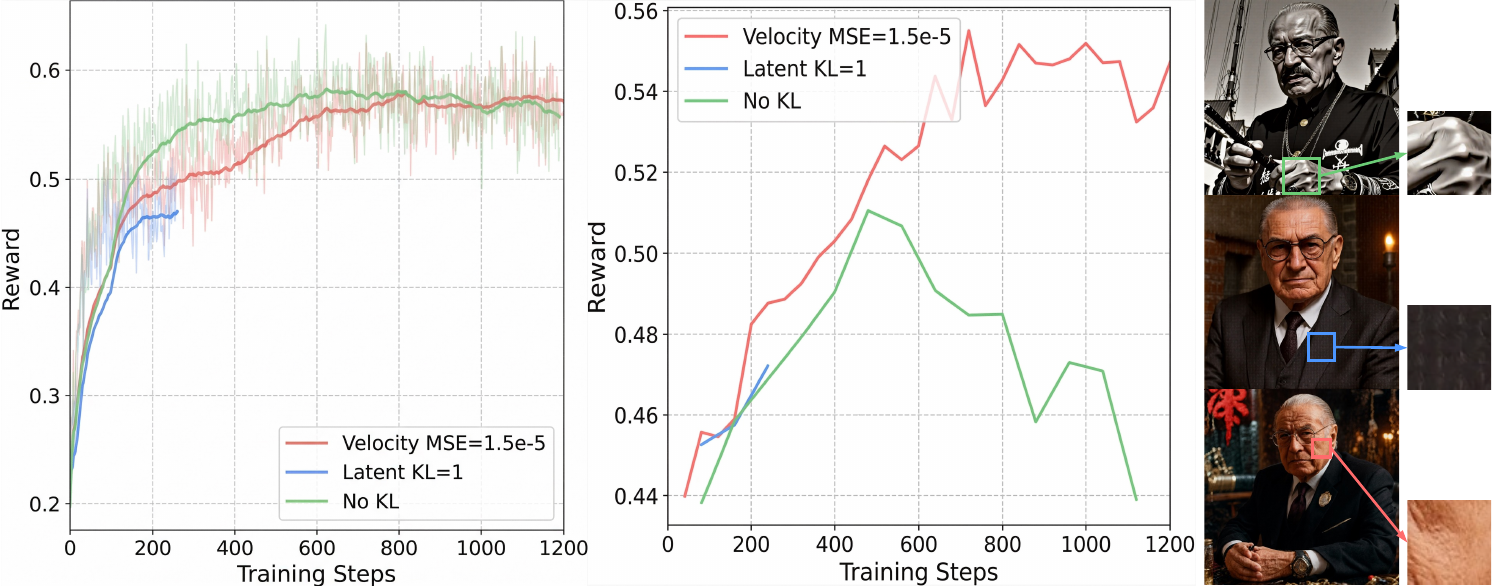}
\caption{\textbf{Ablation Study on Regularization Strategies.} From left to right: training reward, validation reward, and images generated under three different regularization strategies. \textcolor{pltnokl}{Without regularization}, the validation reward drops after an initial increase, leading to unnatural, oversaturated textures in the generated images. For \textcolor{pltkl}{KL divergence on the latents}, the significant drop in training reward indicates that a sufficiently large KL coefficient has been used, yet grid-like artifacts still emerge as early as step 250, prompting us to terminate this run early. In contrast, directly applying \textcolor{pltmse}{MSE regularization} on the velocity field ensures stable training dynamics and produces high-fidelity images with realistic textures.}
\vspace{-2mm}
\label{fig:ablattion_kl}
\end{figure}

\section{Conclusion and Future Work}
\label{sec:conclusion}

In this work, we presented UniGRPO, a unified reinforcement learning framework designed to align interleaved text-and-image generation models. By formulating the multimodal generation process as an MDP, we successfully integrated autoregressive reasoning and flow-based visual synthesis into a single optimization loop. Our minimalist approach establishes a scalable training recipe by eliminating CFG to enforce linear rollouts and employing velocity-based regularization to mitigate reward hacking. Empirically, we demonstrated that UniGRPO effectively enhances image generation quality through chain-of-thought reasoning, providing a scalable baseline for future post-training of fully interleaved models. We identify two promising directions to extend this framework:

\paragraph{Scaling to Multi-Round Interleaved Generation.}
While this work validates UniGRPO on single-round generation (Prompt $\to$ Reasoning $\to$ Image), the unified MDP formulation naturally extends to multi-turn interactions. Future work will apply UniGRPO to complex scenarios such as interactive image editing, visual storytelling, and multi-turn dialogue. In these settings, the model must maintain context consistency across long horizons, where our CFG-free, linear rollout strategy is expected to offer significant computational advantages and training stability.

\begin{figure}[!ht]
\centering
\includegraphics[width=0.9\textwidth]{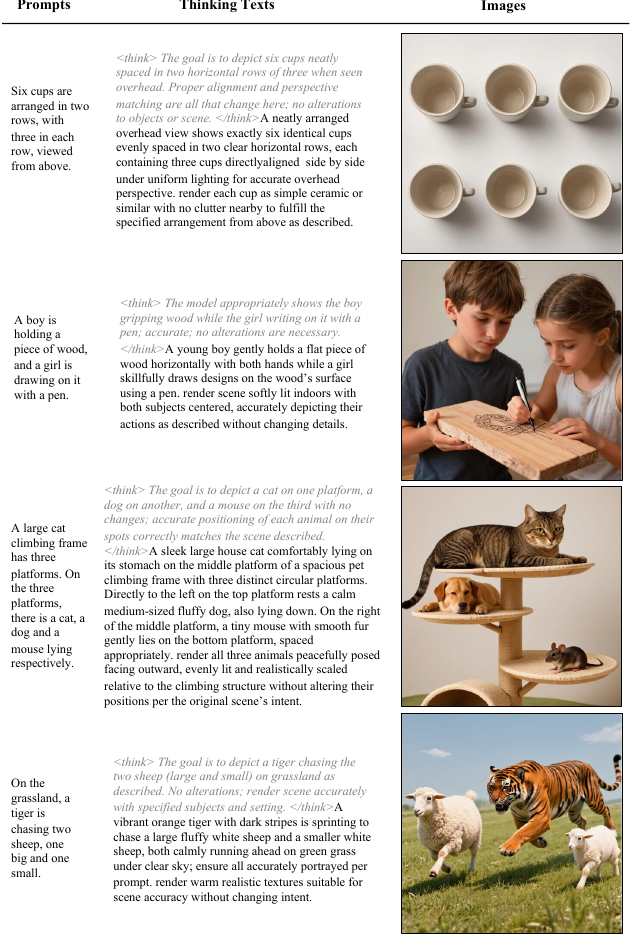}
\vspace{-2mm}
\caption{Reasoning and visual outputs of UniGRPO. Our joint RL optimization produces task-oriented reasoning that guides the synthesis policy toward faithful, photorealistic image generation.}
\label{fig:unigrpo_think_img}
\end{figure}
\vspace{-2mm}
\paragraph{Multimodal Process Reward Modeling.}
Currently, UniGRPO relies on sparse terminal rewards evaluated solely on the final generated image. This can lead to inefficient credit assignment, where the model may generate high-quality images despite flawed reasoning. A critical future direction is to introduce \textit{Multimodal Process Reward Models} (PRMs) that provide dense feedback on the intermediate reasoning steps. By verifying whether the generated "thoughts" are logically sound and aligned with the visual intent before the image is generated, we can further improve the sample efficiency of RL training and ensure better interpretability of the model's decision-making process.

\clearpage

\bibliographystyle{unsrtnat}
\bibliography{
references/flowpo,
references/alignment, 
references/search,
references/proxy_tuning,
references/models,
references/benchmarks,
references/others,
references/unigrpo
}

\clearpage

\appendix
\section{Extended Experimental Results}\label{app:sec:extened-exp-results}

\subsection{Detailed GenEval Results}\label{app:subsec:detailed-geneval}

Table \ref{app:tab:geneval} provides the comprehensive category-level breakdown of the GenEval benchmark. We present the fine-grained metrics across all six sub-categories (Single Object, Two Objects, Counting, Colors, Position, and Attribute Binding) for all evaluated models and baselines.

\begin{table*}[h]
\centering
\caption{Main quantitative results on the GenEval benchmark. All RL methods use the Bagel checkpoint after SFT. ``Thinking'' denotes whether the method explicitly generates intermediate reasoning tokens. $\times$ indicates training collapse.}
\vspace{-2mm}
\label{app:tab:geneval}
\resizebox{\textwidth}{!}{
\begin{tabular}{l c| c c c c c c c}
\toprule
\textbf{Model / Method} & \textbf{Thinking} & \textbf{Overall} & \textbf{Single Obj.} & \textbf{Two Obj.} & \textbf{Counting} & \textbf{Colors} & \textbf{Position} & \textbf{Attr. Binding} \\
\midrule
Bagel & $\times$ & 0.78 & 0.98 & 0.96 & 0.78 & 0.84 & 0.52 & 0.58 \\
Bagel & $\checkmark$ & 0.79 & 0.99 & 0.92 & 0.77 & 0.88 & 0.56 & 0.62 \\
\midrule
SFT & $\times$ & 0.83 & 0.99 & 0.95 & 0.83 & 0.89 & 0.58 & 0.75 \\
SFT & $\checkmark$ & 0.82 & 0.98 & 0.93 & 0.63 & 0.91 & 0.68 & 0.79 \\
\midrule
ReFL & $\times$ & 0.85 & 1.00 & 0.97 & 0.86 & 0.92 & 0.57 & 0.81 \\
ReFL & $\checkmark$ & 0.84 & 0.99 & 0.96 & 0.63 & 0.94 & 0.70 & 0.82 \\
FPO & $\times$ & 0.87 & 0.99 & 0.99 & 0.90 & 0.93 & 0.59 & 0.86 \\
FPO & $\checkmark$ & 0.85 & 0.99 & 0.97 & 0.69 & 0.91 & 0.69 & 0.81 \\
FlowGRPO & $\times$ & 0.88 & 0.99 & 0.98 & 0.93 & 0.94 & 0.60 & 0.86 \\
FlowGRPO & $\checkmark$ & 0.86 & 0.99 & 0.96 & 0.76 & 0.90 & 0.71 & 0.84 \\
TextGRPO & $\checkmark$ & 0.88 & 0.99 & 0.96 & 0.87 & 0.91 & 0.76 & 0.84 \\
\midrule
ReFL (w/ Thinking) & $\checkmark$ & 0.83 & 0.99 & 0.94 & 0.64 & 0.92 & 0.70 & 0.81 \\
\begin{tabular}{@{}l@{}}ReFL (w/ Thinking) \\ + TextGRPO\end{tabular} & $\checkmark$ & 0.87 & 0.98 & 0.97 & 0.84 & 0.91 & 0.75 & 0.80 \\
UniFPO & $\checkmark$ & $\times$ & $\times$ & $\times$ & $\times$ & $\times$ & $\times$ & $\times$ \\
\textbf{UniGRPO (Ours)} & $\checkmark$ & \textbf{0.90} & 0.99 & 0.99 & 0.91 & 0.91 & 0.73 & 0.86 \\
\bottomrule
\end{tabular}
}
\end{table*}

\subsection{Qualitative Results Across Training Stages}
\begin{figure}[htbp]
\centering
\includegraphics[width=1.0\textwidth]{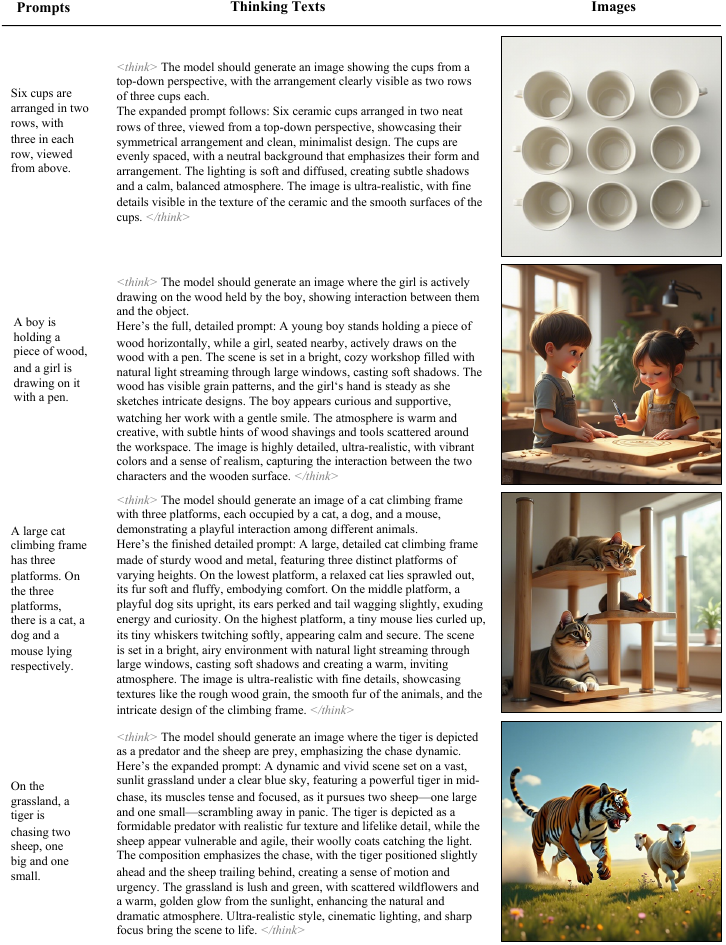}
\caption{Reasoning trace and generated image of the baseline Bagel model.}
\label{fig:bagel_text_img}
\end{figure}

\begin{figure}[!ht]
\centering
\includegraphics[width=1.0\textwidth]{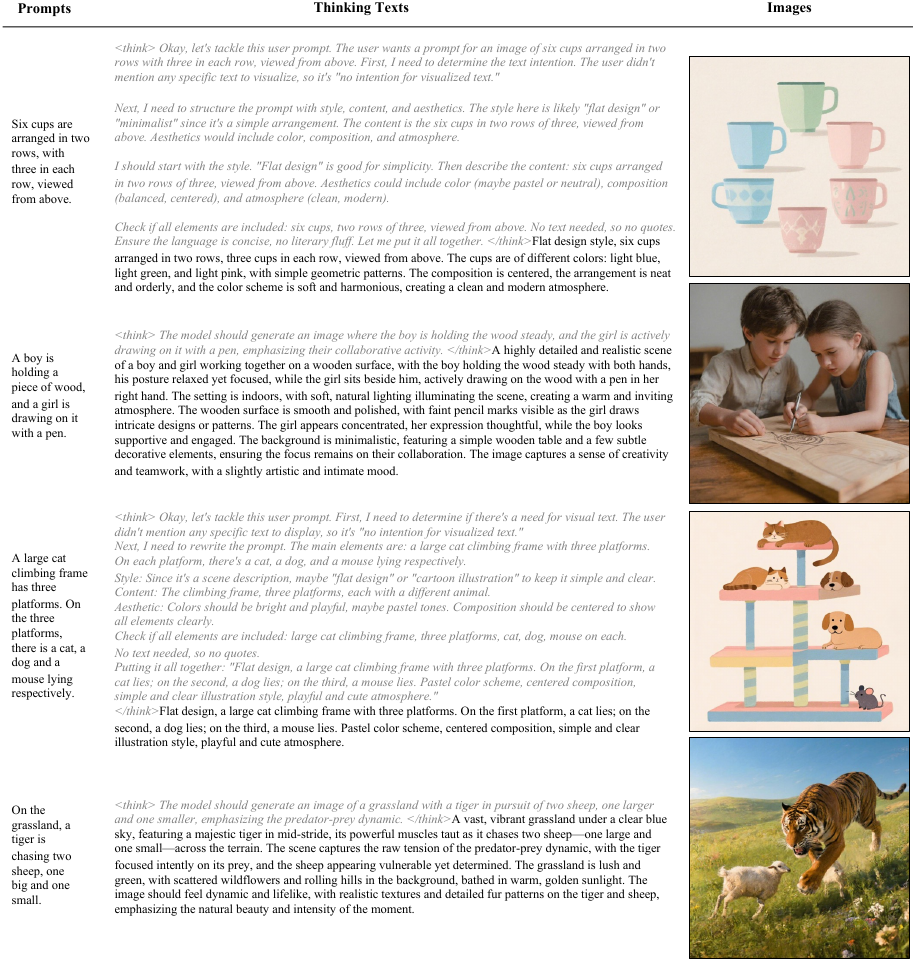}
\caption{Reasoning trace and generation result after SFT. During SFT, we standardized the original \texttt{<think>} format to match prevalent LLM conventions. Although the model produces extensive reasoning steps, the translation from textual thoughts to fine-grained visual details remains suboptimal.}
\label{fig:sft_text_img}
\end{figure}

We present qualitative comparisons of reasoning traces and generated images across different training stages. Figure~\ref{fig:bagel_text_img} shows the reasoning trace and generated image of the baseline Bagel model. Figure~\ref{fig:sft_text_img} illustrates the result after SFT, during which we standardized the original \texttt{<think>} format to match prevalent LLM conventions. Although the SFT model produces extensive reasoning steps, the translation from textual thoughts to fine-grained visual details remains suboptimal. Figure~\ref{fig:unigrpo_think_img} demonstrates that the subsequent RL-based training stage effectively addresses this gap, yielding more coherent reasoning traces and higher-quality image generation.

\section{Implementation Details and Hyperparameters}\label{app:sec:implementation-details}

Table \ref{app:tab:hyperparameters} details the comprehensive hyperparameter settings used for the joint training of our Mixture of Experts (MoT) architecture. This includes the specific optimization configurations for both the text reasoning expert (TextGRPO) and the image denoising expert (FlowGRPO), alongside the joint objective weight $\lambda$.
\begin{table}[h]
\centering
\setlength{\tabcolsep}{9pt}
\small
\caption{Hyperparameters for UniGRPO.}
\begin{tabular}{rl}
\toprule
\multicolumn{2}{c}{\textbf{Model Configuration}} \\
\midrule
Training Timesteps & 25 \\
CFG Scale & 1 \\
Timestep Shift & 3 \\
Image Resolution & 1024 \\
Evaluation Timesteps & 50 \\
\midrule
\multicolumn{2}{c}{\textbf{Training}} \\
\midrule
Group Size & 24 \\
Batch Size & 32 \\
Reasoning Expert Learning Rate & 1e-6 \\
Denoising Expert Learning Rate & 3e-5 \\
PPO Epochs & 2 \\
Flow Objective Weight ($\lambda$) & 1 \\
\midrule
\multicolumn{2}{c}{\textbf{TextGRPO}} \\
\midrule
KL Divergence Loss Weight & 0 \\
Temperature & 1 \\
\midrule
\multicolumn{2}{c}{\textbf{FlowGRPO}} \\
\midrule
MSE Loss Weight & 1.5e-5 \\
Loss Clip Range & 1e-6 \\
SDE Window & [0, 5] \\
SDE Window Size & 3 \\
Noise Level & 0.8 \\
\bottomrule
\end{tabular}
\label{app:tab:hyperparameters}
\end{table}

\end{CJK*}
\end{document}